\documentclass[sigconf,10pt]{acmart}
\usepackage{microtype}
\usepackage{graphicx}
\usepackage{booktabs} 
\usepackage{amsfonts}
\usepackage{amsmath}
\usepackage{listings}
\usepackage{xcolor}
\usepackage{float}
\usepackage{tabularx} 
\usepackage{caption}
\usepackage{subcaption}
\usepackage{adjustbox}
\usepackage{enumitem}
\usepackage{tikz}
\usepackage{pict2e,circledsteps,pgfkeys}
\usepackage{multirow}
\usepackage{geometry}
\usepackage{hyperref}
\usepackage{array}
\usepackage{colortbl}
\usepackage{xurl}
\usepackage{color,soul}
\usepackage{xspace}
\usepackage{comment}
\usepackage{tcolorbox} 
\tcbuselibrary{listings}      

\definecolor{pblue}{RGB}{78,121,167}
\definecolor{pred}{RGB}{225,87,89}
\definecolor{pgreen}{RGB}{89,161,79}
\definecolor{darkgreen}{rgb}{0,0.5,0}

\newtcblisting{examplecode}{
  listing only,
  colback=gray!5,
  colframe=white,
  boxrule=0pt,
  arc=0pt,
  top=-3pt, bottom=-3pt, left=16pt, right=1pt,
  width=\linewidth,
  listing options={
    basicstyle=\ttfamily\small,
    keywordstyle=\color{blue}\bfseries,
    keywords={for,while,pthread_create,tokenize, detokenize,print, alloc_emb, dealloc_emb, alloc_kvpage, dealloc_kvpage, get_next_dist, kv_open, kv_close, kv_fork, kv_remove, kv_close_and_remove, pred, sample, EOS, join_all_threads},
    breaklines=true,
    numbers=left,
    numberstyle=\small,
    numbersep=8pt,
    showstringspaces=false,
    tabsize=2,
    lineskip=-3pt,
    breakindent=10pt,        
    breakautoindent=true   
  }
}

\newcommand{\code}[1]{\texttt{\small #1}}

\newcommand{\paragraphb}[1]{\vspace{0.08in}\noindent{\bf #1.}}

\newcommand{\sys}{Symphony\xspace}
\newcommand{\servingsys}{LLM serving system\xspace}
\newcommand{\servingsyss}{LLM serving systems\xspace}
\newcommand{\lip}{LIP\xspace}
\newcommand{\lips}{LIPs\xspace}

\definecolor{lightgray}{gray}{0.9}
\definecolor{lightblue}{rgb}{0.9,0.9,1}
\definecolor{blue_bg}{rgb}{0.7,0.85,1}
\definecolor{lightyellow}{rgb}{1,1,0.8}
\definecolor{lightpurple}{rgb}{1,0.85,1}
\definecolor{red}{rgb}{1,0,0}
\definecolor{darkgreen}{rgb}{0.4,0.7,0.3}
\definecolor{darkblue}{rgb}{0.2,0.7,0.9}

\AtBeginDocument{%
  }

\copyrightyear{2025}
\acmYear{2025}
\setcopyright{cc}
\setcctype{by}
\acmConference[HOTOS '25]{Workshop on Hot Topics in Operating Systems}{May 14--16, 2025}{Banff, AB, Canada}
\acmBooktitle{Workshop on Hot Topics in Operating Systems (HOTOS '25), May 14--16, 2025, Banff, AB, Canada}
\acmDOI{10.1145/3713082.3730398}
\acmISBN{979-8-4007-1475-7/2025/05}

\begin{document}

\title{Serve Programs, Not Prompts}

\author{In Gim}
\email{in.gim@yale.edu}

\affiliation{%
  \institution{Yale University}
  \city{New Haven}
  \state{CT}
  \country{USA}
}

\author{Lin Zhong}
\email{lin.zhong@yale.edu}

\affiliation{%
  \institution{Yale University}
  \city{New Haven}
  \state{CT}
  \country{USA}
}

\renewcommand{\shortauthors}{Gim et al.}

\begin{abstract}
Current large language model (LLM) serving systems, primarily designed for text completion, are neither efficient nor adaptable for increasingly complex LLM applications due to their inflexible design.
We propose a new LLM serving system architecture that serves \emph{programs} instead of prompts to address this problem. These programs, called \emph{LLM Inference Programs (\lips)}, allow users to customize token prediction and KV cache management at runtime and to offload parts of their application logic, such as tool execution, to the server.
We describe an example of this architecture through a system named \sys, which functions as \emph{an operating system} for \lips. \sys exposes LLM model computations via system calls and virtualizes KV cache with a dedicated file system, while ensuring GPU efficiency with a two-level process scheduling scheme. 
\sys has the potential to open the door to a more efficient and extensible ecosystem for LLM applications.
\end{abstract}





\maketitle

\section{Introduction}

Current large language model (LLM) serving systems are primarily designed for high-throughput text completion.
They accept \emph{prompts} as input and stream generated text as output.
This prompt-serving paradigm has become the de facto standard, commonly available both as cloud APIs and as open-source software~\cite{kwon2023efficient,huggingface_text_generation_inference,sheng2023flexgen}.
Unfortunately, this paradigm faces growing efficiency and adaptability challenges as LLM applications evolve into complex, compound AI systems~\cite{compound-ai-blog}.
For instance, the stateless nature of this paradigm makes multi-round, programmatic interactions with LLMs~\cite{lee2023recursion,yao2023tree,yao2023react} inefficient due to redundant recomputations.
In addition, emerging techniques that deviate from standard token generation process~\cite{holtzman2020the,leviathan2023fast,kuchnik2023validating} or require tool or data-augmented generation workflows~\cite{wang2024executable,jiang2025rago,wang2024survey} can be difficult to implement within these systems, forcing developers to modify the core system~\cite{abhyankar2024infercept,gao2025fast} or create inefficient client-side workarounds (See \S\ref{sec:challenge}).

This paper motivates a new LLM serving architecture that shifts the fundamental \emph{unit of service} from prompts to \emph{programs}. In this model, users flexibly define and offload their own LLM token generation routines to the \servingsys, utilizing fine-grained APIs provided by the system. That is, instead of a prompt, a user sends a program to the serving system to control the generation process.
We term these user-defined routines \emph{LLM Inference Programs (\lips)}. We identify three core properties for APIs to program \lips : (1) decoupling generation from model computation, (2) application-managed model states (e.g., KV cache), and (3) co-locating external interactions. Even fundamental processes like the standard autoregressive generation loop should be explicitly definable using these APIs when needed (See \S\ref{sec:lip}).

To exemplify this idea, we present a new LLM serving system named \emph{\sys} (\S\ref{sec:design}). \sys directly leverages existing operating system abstractions to design the APIs needed to program \lips, and serve them efficiently. 
For example, \sys uses a file system to virtualize the KV cache and a system call for model computation. Users can employ host OS APIs (e.g., POSIX) for custom inference strategies, such as parallel generation or integrating external tools.

The proposed program-serving paradigm enables developers to incorporate application-specific optimizations directly at the LLM generation level and implement new LLM techniques without altering the core LLM serving system. Our initial experiments indicate that \sys can deliver substantial performance improvements, such as achieving up to 7 times greater throughput compared to existing systems like vLLM~\cite{kwon2023efficient}, by empowering users to implement custom KV cache replacement policies through \lips (See \S\ref{sec:result}).

We discuss the main challenges and opportunities of this new approach in \S\ref{sec:discussion}, regarding the granularity of the APIs, security implications, performance overhead, and the need for new benchmarks.

\section{Motivation}
\label{sec:challenge}

Prompt-centric serving systems struggle with efficiency, control, and adaptability when handling complex LLM workflows beyond basic text completion. To address these challenges, developers frequently resort to ad-hoc system modifications or inefficient client-side solutions. Consider the example of developing an LLM-based code editor which provides live code autocompletions.

\begin{itemize}[leftmargin=*]
    \item \textbf{Efficiency:} As the user types, each keystroke ideally triggers an update. A naive prompt-based system recomputes the entire prompt repeatedly. Even with server-side prompt caching~\cite{kwon2023efficient,gim2024prompt}, the policy is server-defined and not application aware. For example, the serving system might cache prompts that will not be used any longer.
    \item \textbf{Interaction:} Suppose the editor implements Retrieval-Augmented Generation (RAG) \cite{khattab2022demonstrate} to fetch relevant API documentation via function calling \cite{openai_function_calling}. This means the client application acts as an intermediary: Prompt $\rightarrow$ Serving System $\rightarrow$ Function Call Spec $\rightarrow$ Client $\rightarrow$ Execute Function (e.g., fetching API documentation) $\rightarrow$ Client $\rightarrow$ Function Result $\rightarrow$ Serving System. Each arrow involves boundary crossing overhead.
    \item \textbf{Control:} Enforcing code conventions or ensuring generated code fits a specific structure (constrained decoding~\cite{beurer-kellner2024guiding}) requires manipulating the token generation process. Prompt-based APIs offer limited control (e.g., temperature, basic JSON mode enforcement), insufficient for complex grammars or custom sampling strategies.
\end{itemize}
We elaborate on these underlying problems below.

\subsection{Resource Inefficiency}

Existing systems suffer from resource inefficiency due to the lack of application-driven management over the KV cache. The KV cache is a crucial component in the efficient serving of ``GPT-style'' LLMs~\cite{radford2018improving,pope2023efficiently}. It enables incremental LLM model computations by avoiding the need to recompute all input tokens. This is possible because the internal representations of a token (K and V states in self-attention) depend solely on preceding tokens in causal Transformers, allowing them to be reused for subsequent LLM model computations when the preceding token sequences remain unchanged.
Optimizing the management and reuse of the KV cache is one of the most critical areas for enhancing LLM serving performance~\cite{kwon2023efficient,gim2024prompt,zheng2024sglang,liu2024cachegen}. 

However, current systems are designed around prompts and lack awareness of application-specific reuse patterns, which limits optimization potential. Typically, KV cache management is governed by a system-wide policy (e.g., LRU eviction) that applies to all requests. For instance, in scenarios involving multi-round prompting, maintaining the KV cache from prior interactions can significantly decrease latency~\cite{gao2025fast}. However, users lack the ability to manage the KV cache retention, even when they possess knowledge of reuse patterns. Current methods, like automatic prefix caching in vLLM, fail to provide this level of flexibility.

Several LLM providers, including Anthropic, offer APIs for prompt caching~\cite{anthropic_prompt_caching}, enabling users to determine what should be cached prior to generation. Systems like SGLang and PromptCache~\cite{gim2024prompt} further allow users to define the structure of input prompts~\cite{zheng2024sglang}, enhancing the efficient reuse of KV caches in parallel generation strategies such as Tree-of-Thought~\cite{yao2023tree}. However, while these systems are beneficial in various contexts, they still handle KV cache management implicitly and lack the capability to accommodate application-specific reuse patterns beyond their intended design, such as using graph~\cite{besta2024graph} or recursive~\cite{lee2023recursion} generation strategies.

\paragraphb{Our solution} We propose making KV cache management an explicit, application-defined operation—shifting optimization responsibility from the serving system to user programs.

\subsection{Communication Overheads}
Current serving systems are only responsible for token generation, forcing any additional non-token generation logic, such as function calling, to be implemented on the client side. This introduces communication overheads. LLM function calling~\cite{patil2024gorilla,yao2023react,openai_function_calling} is essential for interfacing an LLM with external APIs and data sources, which is crucial for many LLM applications~\cite{wang2024survey}. The typical workflow for LLM function calling involves: (1) the user providing a description of functions and their parameters as part of the prompt, (2) the LLM generating a function call in response, and (3) the user parsing and executing the function call, followed by a request to return the execution result to the LLM.

In this process, the user effectively acts as a code interpreter, necessitating network round trips between the user and the \servingsys. These round trips can significantly increase the end-to-end latency of an AI application, especially as the number of function calls increases~\cite{kim2023llm,gim2024asynchronous}. However, in many cases where function calls do not rely on the user's environment for execution (e.g., accessing third-party APIs like weather or stock price APIs, or executing simple code snippets such as NumPy calculations), these round trips can be avoided by enabling the \servingsys to execute the function directly.
Moreover, multi-agent LLM applications~\cite{wu2023autogen} implement inter-agent communication through LLM function calling. However, this agent-to-agent communication incurs high costs because users must handle the communication logic. We can reduce this overhead by enabling the serving system to manage the communication autonomously.

\paragraphb{Our solution} We propose that the serving system should incorporate a code execution environment to handle LLM function calls and code executions internally, rather than relying on the user to manage all execution and communication.

\subsection{Uncontrollable Generation}

Current \servingsyss integrate the autoregressive token generation loop into their core, continuously sampling the next token until they encounter an end-of-sequence (\code{EOS}) token. As a result, users have limited control over token generation, typically restricted to adjusting a few parameters in the next-token sampler and the temperature. This limitation complicates the implementation of emerging techniques for more robust and sophisticated LLM use, such as constrained generation, which ensures the LLM output adheres to a specific format or grammar~\cite{geng2023grammar,beurer-kellner2024guiding}, and policy-based generation~\cite{gupta2024language,kumar2023certifying,kirchenbauer2023watermark} for improved output quality.
These stateful sampling strategies often necessitate intrusive modifications to the \servingsys and expose specialized APIs. While some serving systems support popular methods like constrained decoding through JSON, Regex, and Context-Free Grammar~\cite{dong2024xgrammar,outlines2025,beurer2023prompting}, these methods do not extend to arbitrary sampling strategies.

We note that exposing the sampling process as an end-user facing API is impractical due to the large size of the next-token distribution. For example, GPT-4's vocabulary size exceeds 100K tokens, resulting in a distribution size of approximately 200 KB using FP16.

\paragraphb{Our solution} We propose offloading arbitrary programs to the serving system, enabling applications to directly access and manipulate the token distribution during generation.

\subsection{API Fragmentation}
While ad-hoc solutions can address the limitations of current \servingsys, they often struggle with adaptability as LLM workloads grow more varied and complex. For instance, an application focused on solving complex mathematical problems might leverage parallel reasoning techniques without prioritizing latency~\cite{trinh2024solving}. Conversely, a robotics application with numerous LLM function calls might prioritize reduced latency over absolute accuracy~\cite{chen2023typefly}.

These diverse workloads present challenges in API design, as each specialized solution demands unique API requirements. This has led to a fragmentation of LLM APIs among major providers. For example, Google Cloud~\cite{google_context_caching}, OpenAI~\cite{openai_prompt_caching}, and Anthropic~\cite{anthropic_prompt_caching} each offer distinct API designs and semantics for prompt caching, LLM function calling, and constrained decoding.

\paragraphb{Our solution} We emphasize the need for composable, fine-grained APIs that can accommodate the programming of varied LLM workloads.

\subsection{Related Work}
\label{sec:related}
Recent research aims to make LLM serving systems more application-aware, but it often addresses challenges in isolation without a unified approach. Systems like PromptCache~\cite{gim2024prompt}, SGLang~\cite{zheng2024sglang}, and Parrot~\cite{lin2024parrot} offer APIs for programmatically defining input prompt structures. These systems leverage this structure for efficient KV cache reuse. However, the KV cache management is implicit and cannot express arbitrary reuse patterns beyond the system's predefined abstractions.

For controlled generation, tools such as XGrammar~\cite{dong2024xgrammar}, Outlines~\cite{outlines2025}, and Guidance~\cite{guidance2025} allow users to enforce output constraints using custom rules or domain-specific languages (DSLs). These systems embed a fixed set of decoding strategies directly into the serving stack, limiting extensibility. \sys generalizes this model by exposing the low-level token sampling loop as a programmable interface, enabling arbitrary control strategies that go beyond what built-in grammars or templates can support.

For LLM function calling, InferCept~\cite{abhyankar2024infercept} optimizes the KV cache during such function calls. Some other works, such as LLMCompiler~\cite{kim2023llm} or AsyncLM~\cite{gim2024asynchronous}, make this interaction efficient by allowing multiple function calls to be run concurrently. \sys makes such workflows native, treating function execution, caching, and token generation as composable building blocks within a single user-defined \lips.

\begin{figure}[t]
    \centering
    \includegraphics[width=0.48\textwidth]{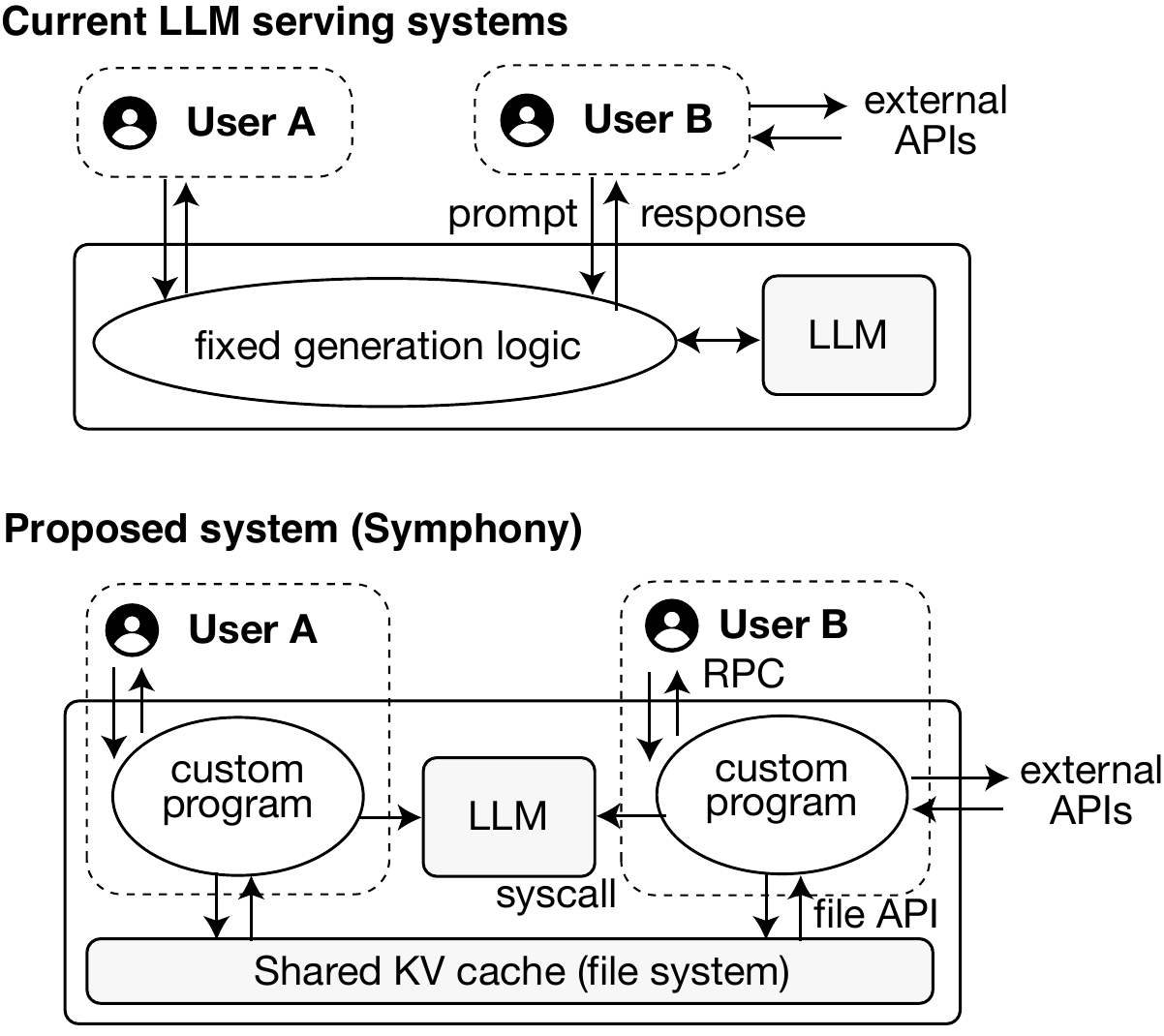}
    \caption{Comparison with existing serving systems (top) and \sys (bottom). \sys serves as an operating system for user-defined inference programs.}
    \label{fig:cpu}
\end{figure}
\section{Program as the Unit of Service}
\label{sec:lip}

In light of the challenges outlined above, we propose that \servingsys should transition from processing mere prompts to handling \emph{programs}, which we term LLM Inference Programs (\lips).
We specifically champion three core attributes for \lips, aimed at empowering users to program their own optimization and generation strategies by utilizing fine-grained APIs for model computation and KV cache management.

\begin{itemize}[leftmargin=1em, itemsep=0mm, topsep=0.2em]
\item \textit{Separation of Generation and Model Computation}:~~
The logic for generating tokens, such as an autoregressive loop, should be decoupled from the model computation, like Transformer model operations on GPUs. The generation process should be defined within \lips, while the \servingsys only needs to focus on efficiently managing the model computations requests.

\item \textit{Application-Controlled Model States}:~~ 
LLM states, including the KV cache, should be able to persist beyond a single \lip and be explicitly managed by the program. It is the user's responsibility, not the \servingsys's, to efficiently utilize the KV cache for their tasks. The \servingsys should offer an API that allows users to manage the KV cache, such as creating, updating, or deleting it.

\item \textit{Integrated External Interactions}:~~ 
Each \lip should independently manage its external interactions, such as function calls and I/O, without depending on the \servingsys or external client-side logic for coordination.
\end{itemize}

\noindent We provide an example code for \lip in \autoref{fig:example}. We elaborate on the API design in the following section.

\section{LLM Serving System as OS}
\label{sec:design}

We introduce a \servingsys called \emph{\sys}, designed to serve \lips. \sys treats the execution of a \lip like a process in an operating system (OS), leveraging existing OS APIs, abstractions, and implementations.
By doing so, we aim to extend the core functions of the \servingsys beyond next token prediction, to include resource virtualization (e.g., KV cache), concurrent program execution, and I/O management—key responsibilities of an OS. 

\begin{figure}[t]
\centering
\begin{examplecode}
// load precomputed kv cache
prefix_kv = kv_open("sys_msg.kv");
suffixes = {
    tokenize(/** query 1 */), ...
    tokenize(/** query n */)
};
for (suffix : suffixes) {
    // fork prefix kv and thread
    kv = kv_fork(prefix_kv);
    pthread_create({
        pos = prefix_kv.len(), step = 0;
        t = suffix;
        // generate until eos token
        while (t != EOS) {
            d = pred(kv, t, pos + step);
            t = sample(d);
            step++;
            print(detokenize(t));
        }
        kv_remove(kv);
    });
}
join_all_threads();
kv_close(prefix_kv);
\end{examplecode}
\caption{Example program demonstrating parallel token generation with shared prefix KV cache.}
\label{fig:example}
\end{figure}

\subsection{Model Computation via System Calls}
\label{sec:pred}
\begin{figure*}[ht!]
    \centering
        \vspace{-1.5em}
    \includegraphics[width=\textwidth]{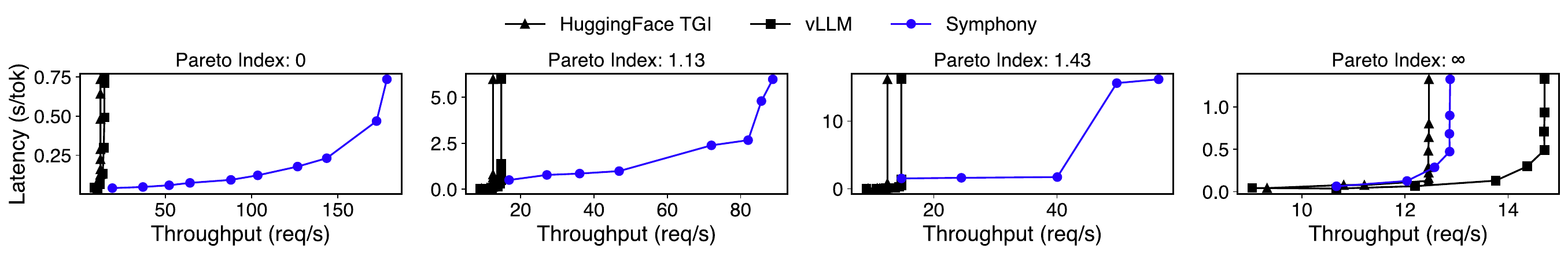}
    \vspace{-1.5em}
    \caption{Estimated performance benefits of prompt caching implemented via \lips in \sys. The figure shows normalized mean end-to-end latency per generated token and throughput using the Llama~\cite{dubey2024llama} 13B model on NVIDIA A100 GPU. \sys enables application-specific LLM optimizations, such as caching frequently reused KV cache, without requiring modifications to the serving system design.}
    \label{fig:gpu}
\end{figure*}

\sys provides a specialized system call for LLM model computation, named \texttt{pred}, which stands for next token prediction. Its signature is as follows:
\begin{verbatim}
pred(kv: kv_file, tokens, positions) -> list[dist]
\end{verbatim}

The \texttt{pred} function accepts two parameters: \texttt{kv}, a pointer to the KV cache file (further details in \S4.3), and \texttt{tokens}, a list of tuples where each tuple contains a token ID and its absolute position within the context. Upon completion of the system call, the KV cache file is updated with new tensors corresponding to the provided \texttt{tokens}, and the function returns a list of next token distributions for each input token.

The \texttt{dist} contains a list of floating-point numbers. With full access to the token distribution, \lips implement various decoding algorithms, such as constrained decoding and speculative decoding. For example, to achieve constrained decoding, \lips integrate a state machine into the generation loop to restrict the distribution variables to those that align with the state machine. For speculative decoding, \lips pass multiple input tokens (draft tokens) to the \texttt{pred} system call and verify them by inspecting the distributions of the tokens.

\subsection{KV Cache Management via File System}
\sys treats the KV cache as files, enabling it to persist beyond a single process's lifecycle, share across multiple processes, and allow \lips to dynamically manipulate it. \sys introduces a specialized file system called \emph{KVFS}. KVFS, similar to traditional file systems, lets \lips manage files (KV cache) using file APIs similar to POSIX, supplemented by specialized APIs for operations like \texttt{extract} and \texttt{merge}. KVFS virtualizes GPU memory areas that store token-level KV tensors in pages, using PagedAttention~\cite{kwon2023efficient}.

In a simple text completion scenario, \lips start by creating an empty file, retrieve its handle with \texttt{kv\_open}, and use this handle with the prompt in \texttt{pred} (See \S\ref{sec:pred}), which then fills the file with the KV cache. They use the same file in the subsequent autoregressive generation loop. For parallel generation with shared prefixes, \lips clone the prefix file for each thread (See \S\ref{sec:thread}), allowing different tokens to fill in without duplicating the actual tensors behind the KV cache.

\lips can directly manipulate files, enabling them to create new files from existing ones by extracting specific token indices with \texttt{extract} or merging existing files into one. This capability benefits inference speedup techniques like runtime context prunning~\cite{xiao2024efficient,tang2022ast}, by removing invalid or unimportant tokens from files. When writing files, \lips can apply a file lock to ensure exclusive access. KVFS enforces access control, allowing only authorized parties to access files. For example, a file containing ``system prompts'' might be readable by all \lips but writable only by the admin.

\subsection{Generations as Threads}
\label{sec:thread}
To support advanced reasoning strategies that leverage parallel generation, \lips spawn multiple threads using POSIX thread APIs. For example, a single \lip implements the entire Tree-of-Thought~\cite{yao2023tree} reasoning, with each thread generating one branch of hypotheses, forking recursively if needed, and joining if a generated hypothesis proves unlikely.

\lips manage I/O independently, eliminating the communication overhead of server-user roundtrips. For instance, \lips initiate LLM function calls by themselves or incorporate arbitrary computation with the generation process. When I/O with external APIs blocks thread execution, \sys interrupts to put the thread in a waiting state. 
\sys exploits this for resource efficiency: when threads wait for I/O, \sys offloads their KV caches from the GPU to the CPU and restores them upon I/O completion. 
Additionally, \lips communicate directly with each other using inter-process communication, which is useful for implementing cooperative multi-agent systems.

\subsection{Two-level Scheduling}

\sys implements a two-level scheduling scheme with two distinct schedulers: the thread scheduler and the batch inference scheduler. The thread scheduler handles CPU scheduling tasks typical of operating systems and executes the thread. When a thread triggers the \texttt{pred} system call for LLM inference, \sys transfers the thread to the ``inference pool'' and marks it as blocked.

Conversely, the inference scheduler aggregates multiple \texttt{pred} system calls into a single batch and schedules this batch on the GPU(s).
Efficient handling of the \texttt{pred} system call by the GPU(s) necessitates batch processing to optimize GPU utilization. Consequently, \sys must strategically batch these calls. The primary challenge lies in timing the batch execution to achieve peak GPU efficiency. Executing the batch prematurely can result in underutilized GPU resources, diminishing throughput, whereas delaying it excessively can increase wait times for the threads. 
\sys dynamically adjusts batch size according to the average frequency of system calls, leveraging models like Poisson process.

\section{Preliminary Results}
\label{sec:result}

We conduct simulated experiments to demonstrate how \sys enables scalable LLM workloads by allowing users to define custom optimization strategies through \lips. We compare \sys with two popular prompt-serving systems, vLLM~\cite{kwon2023efficient} and TGI~\cite{huggingface_text_generation_inference}, in a retrieval-augmented generation (RAG) application scenario. The application inputs a topic, fetches the relevant document, and generates an answer. There are 100 documents, each containing 3,000 tokens.

In this setup, a \lip implements prompt caching~\cite{gim2024prompt} by retaining the KV cache for the top 20 most popular topics and discarding it for others. We evaluate throughput and latency under varying request loads and Pareto indices, which model the skewness of the popularity distribution. The results, shown in~\autoref{fig:gpu}, indicate that \sys outperforms vLLM and TGI when the Pareto index is small (i.e., when a few topics are queried frequently). This improvement arises from \sys's ability to leverage increased cache hits through user-controlled KV cache management. Furthermore, developers can refine \lip logic—e.g., caching only after consecutive requests for the same topic—to optimize performance for specific workload patterns.

We note that these experiments do not yet fully capture the overheads of \sys. Specifically, the current implementation has a simplified design where all \lips are the same and start and finish at the same time; therefore, it does not reflect the scheduling overhead and opportunity costs of using different sampling strategies. We will evaluate these aspects in future work.

\section{Discussion}
\label{sec:discussion}
We believe that bringing programmability to LLM serving systems is a key step towards creating more efficient and capable AI applications. \sys is a step towards achieving this goal. 
The transition proposed by \sys—from serving prompts to executing LLM Inference Programs (\lips) opens up interesting questions about the design and operation of future LLM serving infrastructure. We discuss key challenges and research directions below.

\paragraphb{Finding the right interface granularity}
\sys currently provides one system call, \texttt{pred}, for model computation. This treats a single LLM model forward pass as an atomic operation. This simplifies system design and batching but limits the application's ability to control the model computation process. Enabling finer-grained access (e.g., to attention mechanisms~\cite{beltagy2020longformer, zhang2023h2o} or layer outputs~\cite{chen2024ee}) could unlock powerful application-specific optimizations. However, such interfaces increase complexity, risk violating model abstraction boundaries, and complicate efficient batch scheduling. Determining the optimal granularity—balancing application empowerment against system manageability and performance predictability—remains an open question.

\paragraphb{Security implications}
Executing user-provided \lips fundamentally shifts the trust boundary within the serving system. Unlike prompt-based systems where user input is data, \sys takes code, i.e., \lips, as input. This introduces security vulnerabilities, such as resource exhaustion, model confidentiality, LLM jailbreaking, and parameter extraction via distribution analysis.
Practical deployments necessitate robust sandboxing (e.g., WASM, seccomp filters, lightweight VMs), resource accounting, and fine-grained access control to protect the system and other tenants.

\paragraphb{Performance overhead}
Traditional serving systems achieve high performance through centralized control over batching, scheduling, and GPU pipelining. \sys decentralizes control, empowering \lips to manage their generation logic. This potentially sacrifices global optimization opportunities. \sys's batch scheduler operates with less complete information, potentially leading to suboptimal GPU utilization, especially with heterogeneous \lips. Furthermore, decoupling the sampling logic into \lips hinders server-side optimizations such as pipelining sampling with model execution. Designing intelligent scheduling algorithms and potentially new co-design approaches between \lips and the system to mitigate these performance ``opportunity costs'' is a key research challenge.

\paragraphb{Beyond the OS analogy}
Framing \sys in terms of OS concepts helps convey its broader responsibilities, i.e., resource management, concurrency and I/O, using a familiar mental model. However, this analogy serves more as a scaffold than as an implementation prescription.
Alternative abstractions may be equally, if not more, suitable for realizing the core ideas. For example, language runtimes that support coroutines, actors, or async tasks could offer a natural fit for enabling concurrent, stateful generation workflows. Lightweight execution environments such as WASM or secure containers could provide a sandboxed context for running user-defined logic.
Moreover, we can draw inspiration from other domains where systems evolved into programmable platforms by accepting user-supplied logic, such as OS kernels~\cite{bershad1995extensibility,ebpf2024}, networking~\cite{bosshart2014p4}, and databases~\cite{kulkarni2018splinter}. 
These systems show that we can expose carefully scoped programmability at key abstraction boundaries without re-architecting everything from scratch. This raises a question: can existing LLM serving systems be salvaged? In our view, the tight coupling of sampling, caching, and scheduling in today's prompt-centric architectures makes this difficult. We argue that application-level control, spanning model state management, fine-grained control over generation, and tool use, requires a clean break from the existing architecture. In this sense, a redesign like \sys is not just a matter of elegance but one of necessity.

\paragraphb{Evaluation space}
Standard benchmarks for LLM serving focus on per-prompt throughput and latency and are inadequate for evaluating systems like \sys. The benefits of programmability manifest in the context of complex, multi-step application workflows involving state management (KV cache reuse), external interactions (function calls), and custom control flow (parallel reasoning). Meaningful evaluation requires new benchmarks that capture end-to-end performance, resource consumption (GPU, CPU, memory), and responsiveness for realistic application scenarios, moving beyond isolated token generation metrics.

\vspace{-1.5em}
\section*{Acknowledgments}
This work is supported in part by National Science Foundation (NSF) Athena AI Institute
(Award \#2112562) and Yale University. The authors thank the reviewers for their constructive comments.

\bibliographystyle{abbrv}
\bibliography{abr-short,main}

\end{document}